%% file: main.tex
\PassOptionsToPackage{table}{xcolor}

\documentclass[a4paper, 10pt, conference]{ieeeconf}      
\usepackage{FG2024}
\usepackage{times}
\usepackage{epsfig}
\usepackage{graphicx}
\usepackage{amsmath}
\usepackage{amssymb}
\usepackage{multirow}
\usepackage{xcolor}
\usepackage{booktabs}

\usepackage{gensymb}
\usepackage{subfigure}
\usepackage[pagebackref,breaklinks]{hyperref}
\FGfinalcopy 

\IEEEoverridecommandlockouts                              
\def\FGPaperID{6} 

\title{\LARGE \bf
    Gait Recognition from Highly Compressed Videos
}

\author{\parbox{16cm}{\centering
   {\large Andrei Niculae$^{1}$\thanks{$^{1}$andrei.niculae101@gmail.com},
   Andy Catruna$^{2}$\thanks{$^{2}$andy\_eduard.catruna@upb.ro}, 
   Adrian Cosma$^{3}$\thanks{$^{3}$ioan\_adrian.cosma@upb.ro}, 
   Daniel Rosner$^{4}$\thanks{$^{4}$daniel.rosner@upb.ro}, 
   Emilian Radoi$^{5}$\thanks{$^{5}$emilian.radoi@upb.ro}
   }\\
   {\normalsize
   National University of Science and Technology Politehnica Bucharest\\}}%
}

\usepackage{fancyhdr}
\begin{document}

\ifFGfinal
\thispagestyle{empty}
\pagestyle{empty}
\else
\author{Anonymous FG2024 submission\\ Paper ID \FGPaperID \\}
\pagestyle{plain}
\fi
\maketitle


\thispagestyle{fancy}

\begin{abstract}
Surveillance footage represents a valuable resource and opportunities for conducting gait analysis. However, the typical low quality and high noise levels in such footage can severely impact the accuracy of pose estimation algorithms, which are foundational for reliable gait analysis. Existing literature suggests a direct correlation between the efficacy of pose estimation and the subsequent gait analysis results. A common mitigation strategy involves fine-tuning pose estimation models on noisy data to improve robustness. However, this approach may degrade the downstream model's performance on the original high-quality data, leading to a trade-off that is undesirable in practice. We propose a processing pipeline that incorporates a task-targeted artifact correction model specifically designed to pre-process and enhance surveillance footage before pose estimation. Our artifact correction model is optimized to work alongside a state-of-the-art pose estimation network, HRNet, without requiring repeated fine-tuning of the pose estimation model. Furthermore, we propose a simple and robust method for obtaining low quality videos that are annotated with poses in an automatic manner with the purpose of training the artifact correction model. We systematically evaluate the performance of our artifact correction model against a range of noisy surveillance data and demonstrate that our approach not only achieves improved pose estimation on low-quality surveillance footage, but also preserves the integrity of the pose estimation on high resolution footage. Our experiments show a clear enhancement in gait analysis performance, supporting the viability of the proposed method as a superior alternative to direct fine-tuning strategies. Our contributions pave the way for more reliable gait analysis using surveillance data in real-world applications, regardless of data quality.
\end{abstract}

\section{Introduction}
\input{sections/1.introduction}

\section{Related Work}
\input{sections/2.relatedwork}

\section{Method}
\input{sections/3.method}

\section{Results}
\input{sections/4.results}

\section{Conclusions}
\input{sections/5.conclusions}

\section*{Acknowledgements}
\input{sections/6.ack}

{\small
\bibliographystyle{ieee}
\bibliography{refs}
}

\end{document}

%% file: sections/1.introduction.tex
Gait analysis has emerged as a key biometric technique for person identification in video surveillance data, due to its non-invasive nature and the difficulty of obscuring gait patterns. This form of analysis has considerable applications in security and forensic investigations, augmenting systems in scenarios where other biometric recognition approaches may not work due to poor image quality, low resolution or obstructions.

Gait analysis is a powerful tool for understanding various aspects about humans including demographic attributes, emotions, personality traits, and mental health \cite{cosma2023psymo}. However, gait analysis is generally utilized in controlled laboratory settings \cite{CASIA:Yu}, which limits its potential for real-world applicability. Although surveillance footage can be a rich source of gait data for both analysis and pretraining of deep learning models \cite{cosma22gaitformer}, it is often of poor quality and noisy, making it challenging to utilize.

\begin{figure}[hbt!]
    \centering
    \includegraphics[width=\linewidth]{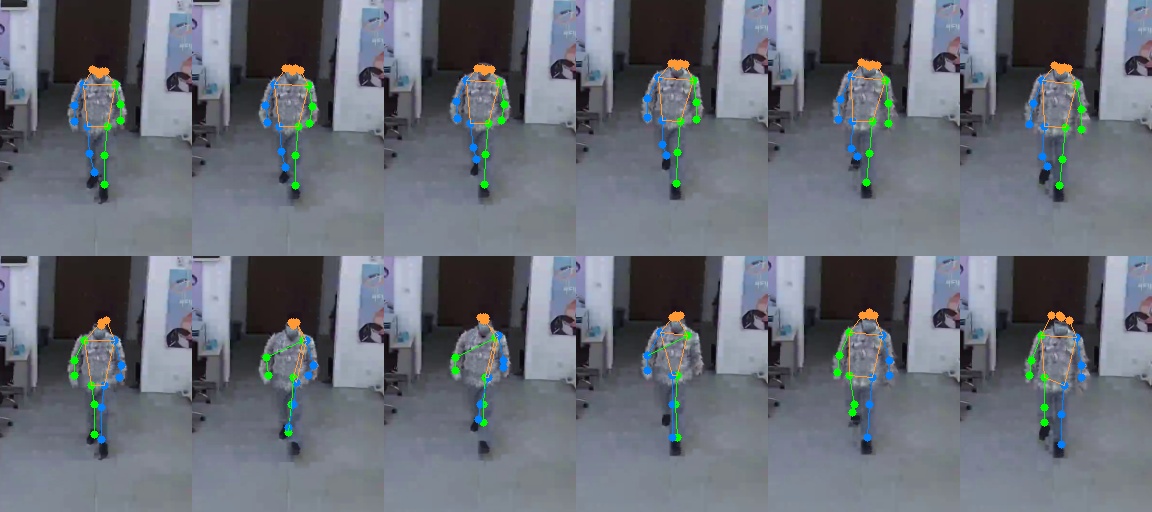}
    \caption{Examples of highly inaccurate pose estimations under severe video degradation (\textbf{second row}), compared to ground truth poses (\textbf{first row}). Artifacts introduced by typical video compression methods hinder the performance of state-of-the-art pose estimation models \cite{sun2019deep}.}
    \label{fig:example-bad-poses}
\end{figure}

Previous works in model-based gait analysis \cite{catruna2023gaitpt, teepe2022towards} have shown that the performance of the pose estimation model has a significant impact on the recognition accuracy. However, most publicly available pose estimation models \cite{xu2022vitpose,alphapose,sun2019deep}, are trained on datasets (e.g. MPII \cite{andriluka14cvpr}, COCO \cite{lin2015microsoft}, OCHuman \cite{zhang2019pose2seg}) of images with high quality and high resolution where the human joints can be accurately detected. Therefore, models exhibit good pose estimation performance on images with similar quality, but fail to generalize to more difficult scenarios such as surveillance footage which suffers from quality degradation. In degraded surveillance footage, pose estimation is particularly challenging due to factors such as low resolution, occlusion, and motion blur. Figure \ref{fig:example-bad-poses} exemplifies this problem. When the videos are highly degraded by typical compression methods (e.g. H.264 compression), even state-of-the-art pose estimation models (e.g. HRNet \cite{sun2019deep}) make erroneous predictions. We show that the performance of the gait analysis systems on low quality images is hindered by the weakened performance of the underlying pose estimation model.  

One straightforward approach to improve the performance of gait analysis on surveillance footage is to fine-tune a pose estimation model on a dataset of images with a similar quality degradation. However, this method is not optimal as it can lead to a loss in performance on the original, undegraded, dataset for downstream applications such as gait analysis. The model may overfit to the noise and artifacts in the surveillance footage \cite{jiang2021towards}, making it specialized for that specific setting but at the cost of catastrophic forgetting \cite{french1999catastrophic}.

To address this issue, we propose incorporating a separate model in the pose estimation pipeline which is trained to correct the artifacts introduced by the quality degradation of surveillance footage. The purpose of this artifact correction model is to modify the input image in a way that improves the accuracy of the pose estimation, while keeping the underlying pose estimation model unaffected. More specifically, the artifact correction model is optimized to minimize the error between the output of the pose estimation model and the ground truth pose.

In this work, we explore the effect of H.264 compression artifacts within surveillance footage and their impact on the performance of pose estimators. We employ HRNet \cite{sun2019deep} and VitPose \cite{xu2022vitpose}, two state-of-the-art pose estimation models for our analysis on the effects of quality degradation. By integrating a fine-tuned artifact correction model (i.e. FBCNN \cite{jiang2021towards}) designed specifically to improve pose estimation performance, we not only refine the performance of the pose estimator in low-quality video scenarios but also preserve its proficiency on uncorrupted data \cite{ehrlich2021analyzing}. Finally, we analyse the performance of gait analysis models on poses obtained with different combinations of image degradation, artifact correction and fine-tuning, demonstrating the advantages of our proposed method. We conduct our analysis on the PsyMo \cite{cosma2023psymo}, a dataset constructed to benchmark gait recognition models in controlled settings under various confounding factors. Compared to other datasets \cite{CASIA:Yu}, PsyMo offers high resolution videos, allowing us to study the effects of video compression, while maintaining high-quality ground truth on the original videos.


Our main contributions are as follows:
\begin{itemize}
    \item We propose an automatic method of obtaining pose estimation training and evaluation data in low quality settings, without the use of any time-consuming manual annotations. We obtain ground truth poses using state-of-the-art models on high quality images and use the poses extracted from degraded training videos with the H.264 compression as training data. 
    
    \item We propose a 2-stage approach for obtaining the poses in highly degraded video footage, that outperforms the typical approach of directly fine-tuning the pose estimation model. Our proposed approach consists of a trainable artifact correction model and a fixed pose estimator. The artifact correction model is designed to improve the performance of downstream pose estimation models by being specifically fine-tuned for this task. We outperform the typical fine-tuning approach both in terms of Average Precision (AP) on the test set (0.956 vs 0.935) and in terms of performance on downstream tasks (40.9\% vs 35.4\% in gait recognition). 
    
    \item We present an analysis of the impact of compression artifacts on the performance of pose estimation models in the context of gait analysis, providing valuable insights into how video quality degradation affects recognition.
\end{itemize}

%% file: sections/2.relatedwork.tex
Gait recognition is an extensively studied problem in biometrics \cite{catruna2023gaitpt,cosma22gaitformer,gpgait2023,fan2020gaitpart}, enabling person recognition without explicit cooperation and is robust to changes in appearance. Typically, gait recognition is performed using either sequences of silhouettes \cite{fan2020gaitpart,chao2019gaitset}, or through processing sequences of poses \cite{catruna2023gaitpt,cosma22gaitformer,gpgait2023} extracted with pose estimation models \cite{xu2022vitpose,sun2019deep}. While silhouettes usually encode some outward appearance features which confound results, pose sequences mostly \cite{cuatrunua2024paradox} encode only movement features. Methods such as GPGait \cite{gpgait2023} and GaitPT \cite{catruna2023gaitpt} employ a hierarchical approach to encode both fine-grained, joint-level, movement as well as high level patterns, at the level of body parts. Multiple datasets have been proposed for gait recognition, the most popular being CASIA-B \cite{CASIA:Yu}, a controlled gait dataset having people walk under various walking registers and multiple viewpoints. However, videos from CASIA-B are of low-resolution (i.e., 320x240), limiting the analyses of the impact of compression on downstream gait recognition tasks. Recently, PsyMo \cite{cosma2023psymo} was proposed as a high resolution drop-in replacement for CASIA-B, having the same controlled setting with a larger amount of walking registers and subjects. 


Multiple previous works have studied the problem of artifact correction under the guidance of downstream tasks \cite{galteri2017deep,katakol2021distributed,jiang2021towards,ehrlich2021analyzing}. For instance, Galteri et al., \cite{galteri2017deep} showed that a GAN model that removes compression artifacts improves object detection tasks. Furthermore, Ehrlich et al., \cite{ehrlich2021analyzing} trained a task-adapted artifact correction model to mitigate JPEG artifacts that hinder the performance of downstream object detection and semantic segmentation models. The authors directly use the supervision signal from the downstream model to train the correction model, while keeping the downstream model frozen. We adapt this framework and apply it to pose estimation, aiming to further improve upon downstream pose-based gait recognition performance. Compared to the work of Ehrlich et al., \cite{ehrlich2021analyzing}, we study the effects of video compression rather than JPEG artifacts on downstream pose estimation tasks. Furthermore, we are the first, as far as we know, that show that task-adapted artifact correction improves gait recognition performance under highly degraded videos.

%% file: sections/3.method.tex
\begin{figure}[hbt!]
    \centering
    \includegraphics[width=0.85\linewidth]{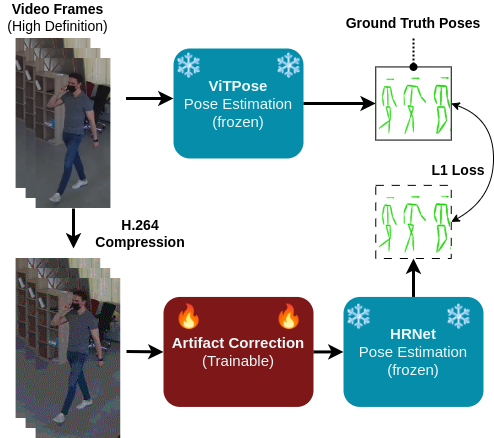}
    \caption{Overall diagram of our method for training an artifact correction model without any manual labels. We utilize a pose estimation model on high definition video frames to obtain robust ground truth poses. The quality of the videos is highly decreased with H.264 compression to simulate real-world environments. We train an artifact correction model to alter the image so that a separate frozen pose estimation model obtains poses close to the ground truth.  }
    \label{fig:diagram}
\end{figure}

\subsection{Automatic Dataset Construction}
To the best of our knowledge, there is no dataset containing low quality videos of people walking along with robust pose estimation annotations. Consequently, we automatically construct a dataset that simulates the setting of surveillance footage without the need for manual labelling. We utilize PsyMo \cite{cosma2023psymo}, a gait analysis dataset that contains high quality videos of over 300 people walking, recorded in constrained settings with multiple cameras and scenarios.

We employ ViTPose-H \cite{xu2022vitpose} \footnote{\href{https://github.com/ViTAE-Transformer/ViTPose}{github.com/ViTAE-Transformer/ViTPose}. Accessed 17.03.2024}, a state-of-the-art model for pose estimation, to obtain the poses for every frame in the PsyMo videos. These predicted poses act as a ground truth in our training scenarios as they were obtained with the top-performing version of the model on high quality and high resolution videos. 

To obtain surveillance-like videos, we utilize the H.264 video compression \cite{10.5555/1942939}, one of the most common video compression algorithms \cite{Abdul2016} to simulate low quality scenarios, using a high compression rate of 45 crf.  As the ground truth poses were computed from high quality images, we obtain pairs of low quality images similar to those found in surveillance videos \footnote{\href{https://www.networkwebcams.co.uk/blog/h264-video-compression-in-ip-video-surveillance-systems/}{https://www.networkwebcams.co.uk/blog/h264-video-compression-in-ip-video-surveillance-systems/}. Accessed 17.03.2024} and their corresponding pose estimation annotations. Obyrne et al. \cite{obyrne2022impact} studied the impact of H.264 compression and showed that high compression rates lead to a notable drop in detection performance, in the case of object detection using a YOLOv5 model. We utilize this resulting dataset to both fine-tune a pose estimation model and a task-adapted video artifact correction module to improve estimation. 

\subsection{Task-Adapted Video Artifact Correction}

The high-level overview of our approach to obtaining robust pose estimations from low quality videos is shown in Figure \ref{fig:diagram}. We obtain ground truth poses by employing ViTPose, a state-of-the-art pose estimation model, on high definition videos. We degrade the quality of the videos by utilizing H.264 compression which introduces artifacts that affect the performance of the pose estimation model. We train a model to correct the artifacts in the low quality videos so that a separate pose estimation model obtains similar poses to those in the ground truth. 

Formally, we consider an initial high quality video of a walking person $V = \{I_1, I_2,..,I_N\}$ where $I_i$ is a video frame and $N$ is the total number of frames in the video. For each high quality video $V$ in the dataset we utilize ViTPose to obtain a sequence of ground truth skeletons $P = \{p_1, p_2,..,p_N\}$ , where $p_i \in R^{17 \times 2}$ corresponds to the pose of the person in the frame $I_i$ We apply H.264 compression to obtain a low quality video $V' = \{I'_1, I'_2,..,I'_N\}$, where $I'_i$ is the degraded correspondent of the $I_i$ video frame.

We utilize HRNet on the low quality videos to obtain the pose predictions $\hat{P} = \{\hat{p}_1, \hat{p}_2,..,\hat{p}_N\}$. The predicted poses $\hat{P}$ are not equal to the ground truth poses $P$ due to the artifacts introduced by the employed video compression. To solve this, we train an additional model that corrects the artifacts in the low quality videos with a training signal that explicitly aims to optimize the performance of the pose estimation model, similar to the approach of Ehrlich et al. \cite{ehrlich2021analyzing}. In this framework, the pose estimation model is frozen, and only the artifact correct module is trainable.

The artifact correction model takes as input a low quality image $I'_i$ and outputs an image $I^*_i$ on which the pose estimation model performs better. We utilize HRNet on these corrected poses to obtain $\hat{P^*} = \{\hat{p}^*_1, \hat{p}^*_2,..,\hat{p}^*_N\}$. The artifact correction model is trained to correct the images so that the poses $\hat{P^*}$ are as close as possible to the ground truth poses $P$. The training loss that we employ can be written as:
    \begin{equation}
        L_{AC} = \sum_{V}^{} \sum_{i=1}^{N} |\hat{p}^*_i - p_i |
    \end{equation}

\subsection{Implementation Details}

For obtaining the ground truth poses we selected the top-performing model on the COCO keypoints estimation benchmark \cite{DBLP:journals/corr/LinMBHPRDZ14}, namely ViTPose, to ensure the most accurate automatic labelling. We chose to train and test a different human pose estimation model (i.e. HRNet) in our experiments to eliminate the bias of utilizing the same architecture and weights for both training and annotation. We employ the FBCNN architecture \cite{jiang2021towards} for correcting the images as it has available pre-trained weights \footnote{\href{https://github.com/jiaxi-jiang/FBCNN/releases/tag/v1.0}{github.com/jiaxi-jiang/FBCNN}. Accessed 17.03.2024}, is trained for a similar task (artifact correction in JPEG compression), and is flexible due to its quality factor hyperparameter. In our experiments we report only results of the fine-tuned version, as we did not observe any improvements when utilizing FBCNN out-of-the-box.

For fine-tuning the artifact correction and pose estimation models, we utilize the official pre-trained weights \footnote{\href{https://github.com/HRNet/HRNet-Human-Pose-Estimation}{github.com/HRNet/HRNet-Human-Pose-Estimation}. Accessed 17.03.2024}. For HRNet, we select the base version of the model which expects an image size of 192x256 pixels.
\begin{figure*}[hbt!]
    \centering
    \includegraphics[width=0.45\textwidth]{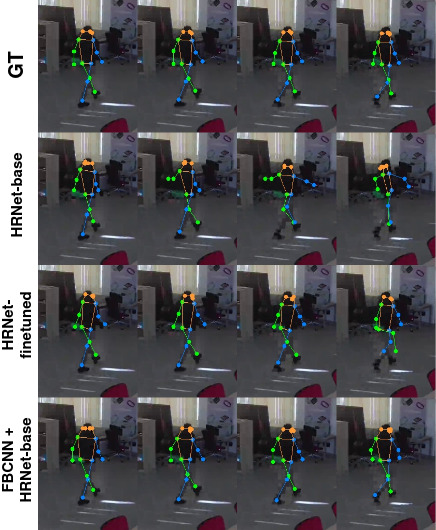}
    \includegraphics[width=0.45\textwidth]{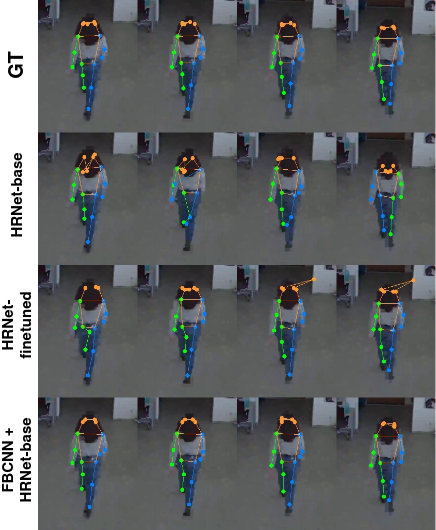}
    \caption{Examples of poses extracted from highly degraded videos with the 3 experimental approaches on the test set. \textbf{First row} - Ground Truth poses; \textbf{Second row} - poses obtained with the original pre-trained HRNet model; \textbf{Third row} - poses obtained with fine-tuned HRNet; \textbf{Fourth row} - poses obtained artifact correction model (FBCNN) in combination with the original HRNet model. The proposed method obtains the most accurate poses, closely resembling those from the ground truth.}
    \label{fig:example-bad-poses-corrections}
\end{figure*}
The fine-tuning  of HRNet is performed with the default configuration of hyperparameters, except for the batch size which is changed to 240. For FBCNN, we modify the training to have the objective of improving pose estimation performance, while keeping the pose estimator frozen. The quality factor is fixed at 0.5 during both training and inference, since we only fine-tune the image reconstruction module of FBCNN using our modified loss function. The videos are compressed with h264 with a value of crf equal to 45. The fine-tuning training time takes 6 hours for HRNet and 6 hours for FBCNN on 3 NVIDIA A100 40GB GPUs.

As we are interested in generating images on which the pose estimation model performs better, we crop the images using the bounding box of the person. We add a padding of 600px vertically and 400px horizontally to the center of the bounding box of the person, as we observed that simply cropping the image according to the bounding box leads to worse performance. We further filter the dataset by eliminating frames that change the ratio of the image when resized to the input size of the HPE model (192x256). The final dataset consists of 984,532 frames, split by identity under the official PsyMo splits \cite{cosma2023psymo}: 80\% train, 10\% validation, 10\% test.

%% file: sections/4.results.tex
\begin{figure*}[hbt!]
\centering
\subfigure[]{\includegraphics[width=0.3\textwidth]{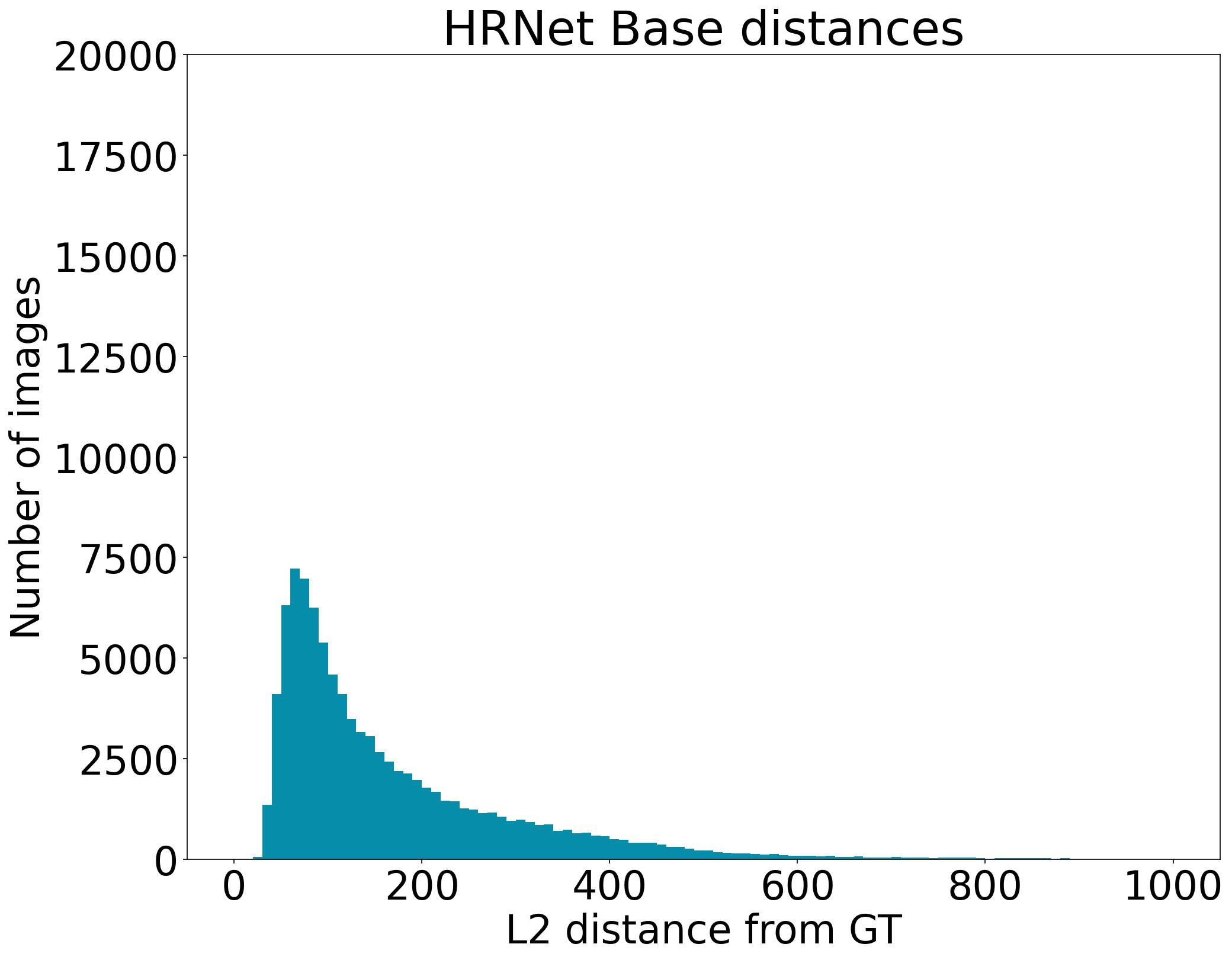}}
\subfigure[]{\includegraphics[width=0.3\textwidth]{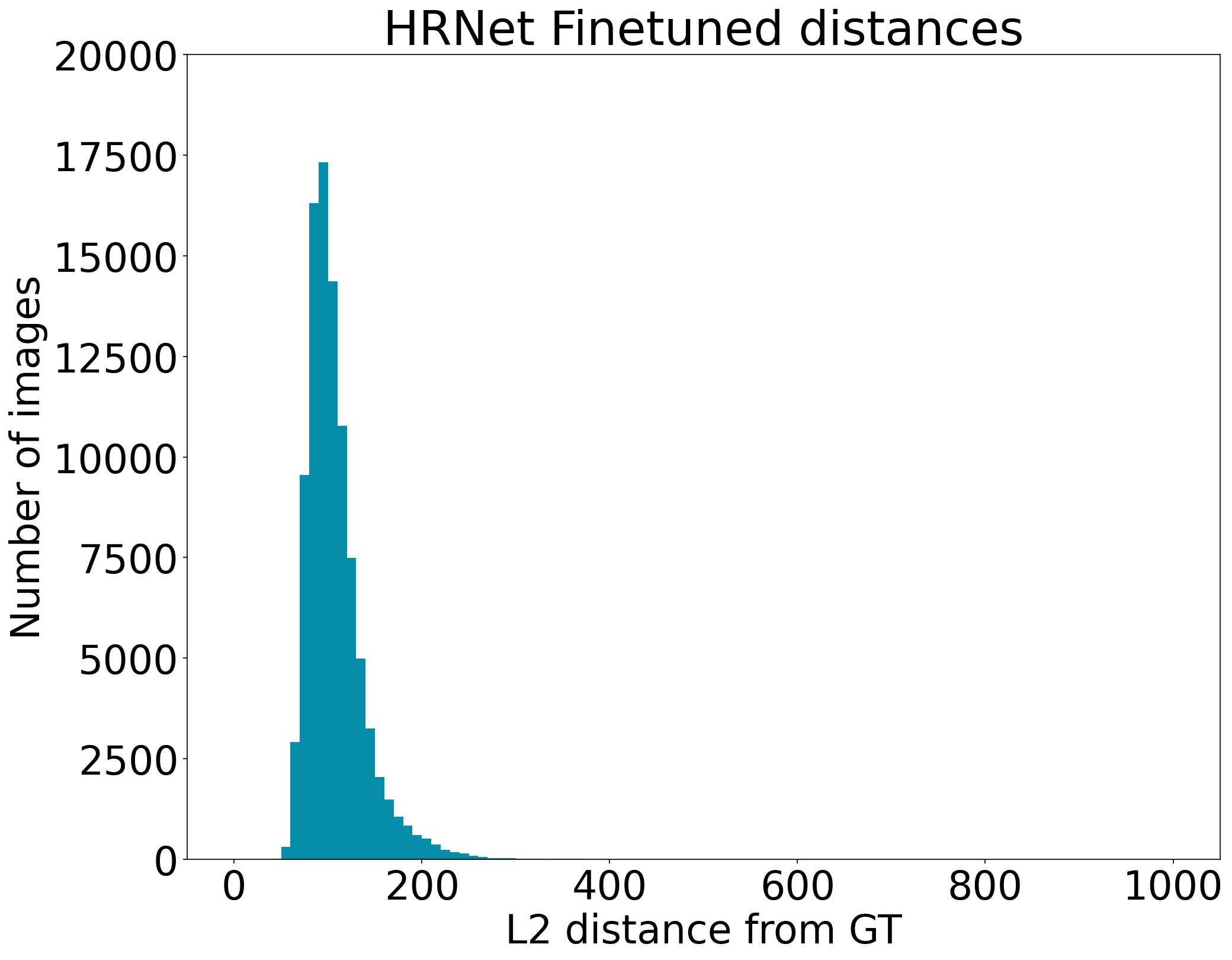}}
\subfigure[]{\includegraphics[width=0.3\textwidth]{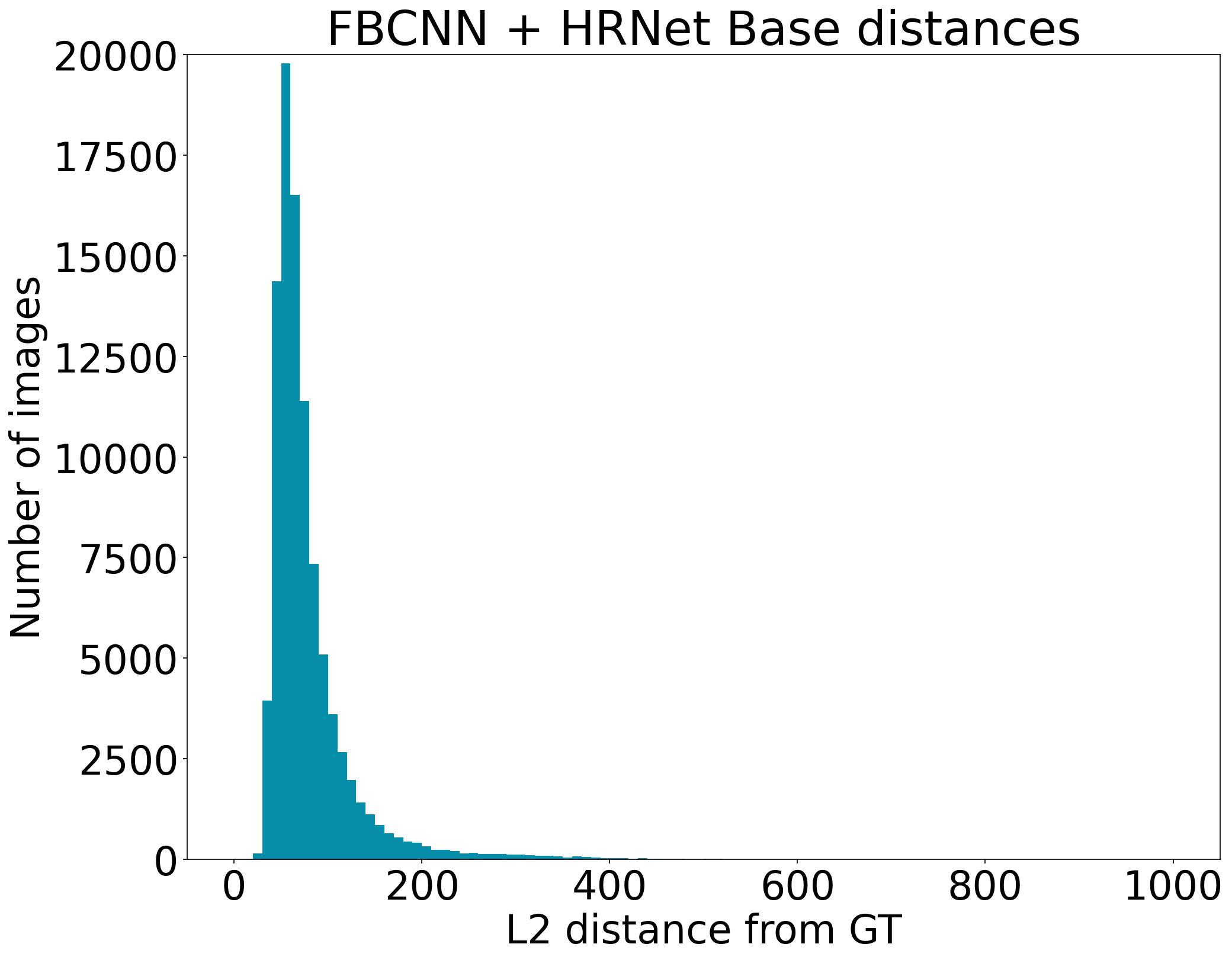}}
\caption{Histograms of L2 distances from ground truth of test-set poses. \textbf{a)} poses obtained with pre-trained HRNet (mean: 173.32); \textbf{b)} poses obtained with finetuned HRNet (mean: 107.53); \textbf{c)} poses obtained with FBCNN + pre-trained HRNet (mean: 79.13)}
\label{fig:dist-histograms}
\end{figure*}

\subsection{Pose Estimation in Low Quality Settings}

We experiment with 3 different methods of obtaining pose estimation in highly degraded videos. The first one involves utilizing a HRNet pose estimation model pre-trained on the COCO dataset. The second method consists of using the same architecture, an HRNet, but fine-tuned on the highly degraded videos with the poses obtained on the original images as ground truth. The third method, as shown in Figure \ref{fig:diagram}, proposes to keep the HRNet frozen and fine-tune a separate artifact correction architecture to adjust the low quality image so that the pose estimation performance is improved.

\input{tables/pose_results}

\begin{figure}[hbt!]
\centering
\includegraphics[width=\linewidth]{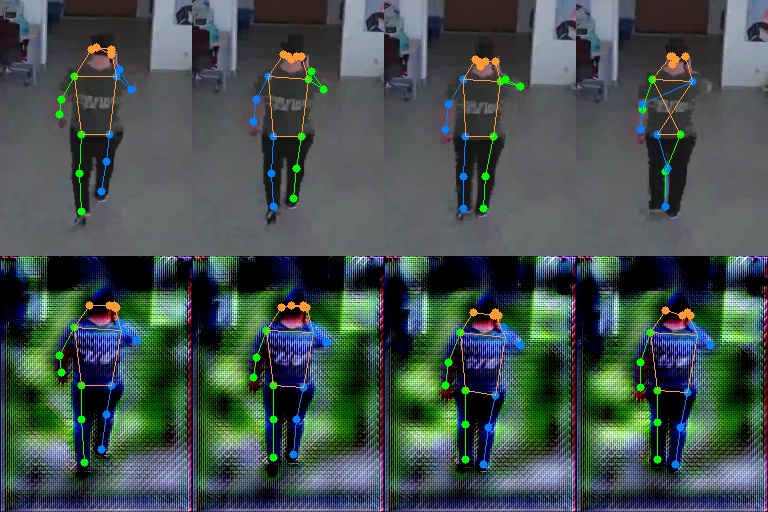}
\caption{Examples of corrections made by the fine-tuned FBCNN model on degraded videos. The model learns to obscure the background with the purpose of better highlighting the human body and its joints.}
\label{fig:example-fbcnn-correction}
\end{figure}

Table \ref{tab:pose_results} shows the results of the 3 pose estimation models on the test set compared to the ground truth poses. The pre-trained version of the HRNet suffers a considerable loss in performance in terms of Average Precision (AP) from 0.935 on the high quality images to only 0.783 on the compressed video frames. This shows that current pose estimation models may not be robust to commonly utilized video compression methods that are known to introduce artifacts.

Fine-tuning the pose estimation models increases the performance of the model on the highly degraded images to 0.935. Furthermore, it also slightly improves the AP on the original high quality images. However, our proposed two-stage method consisting of a fine-tuned artifact correction model in conjunction with a pre-trained pose estimation model obtains the highest Average Precision in both cases.

\input{tables/gait_results_degraded}

In Figure \ref{fig:example-bad-poses-corrections} we show qualitative examples of pose predictions on the test set for all experimental approaches. The ground truth poses displayed in the first row are obtained with the VitPose model on the original high quality images. The poses obtained with a pre-trained pose estimation model on highly degraded videos are shown in the second row. It can be observed that these poses are exceedingly distorted due to the artifacts introduced by compression methods. Fine-tuning the pose estimation model improves the quality of the poses (third row), but the predictions on occluded joints (such as the eyes) can sometimes be incorrect. Finally, the poses obtained with the proposed method (fourth row) are the most robust of the 3 approaches, showcasing the advantages of fine-tuning a separate artifact correction model with pose estimation training signal.

In Figure \ref{fig:example-fbcnn-correction} we showcase images corrected by our trained FBCNN model. Because the fine-tuning objective is task-adapted on pose estimation signal, and not direct artifact correction, the output images are noticeably altered. However, they improve the HPE model predictions, potentially because the artifact correction model learns to obscure the background, highlighting individuals within the image frame and increasing details on the person's joint locations. 

The COCO mAP metric may not fully capture how the fine-tuned models improve the HPE performance. For each corrected pose we also compute the L2 distance between the predicted and ground truth keypoints. The histograms of these distances are displayed in Figure \ref{fig:dist-histograms} showing how each model affects the performance. We observe that the FBCNN + HRNet Base combination has the lowest mean distance, correlating with COCO's mAP.

\subsection{Evaluation on Downstream Gait Recognition}

\input{tables/gait_results_original}

We evaluate the proposed method of obtaining robust pose estimations in low quality settings on the downstream task of gait re-identification. We employ the GaitPT architecture \cite{catruna2023gaitpt}, a state-of-the-art hierarchical transformer designed for gait analysis. We train GaitPT on the high quality poses obtained with VitPose and test the model on 3 different types of poses obtained from low quality videos: a) obtained with pre-trained HRNet; b) obtained with fine-tuned HRNet; c) obtained with fine-tuned FBCNN and pre-trained HRNet.

The evaluation protocol that we employ for gait recognition on PsyMo is similar to that of CASIA-B. We utilize the same data for training the gait recognition model as the data that was utilized for fine-tuning the pose estimation and artifact correction models. In terms of the testing data, we split it into pairs of gallery and probes based on each unique combination of scenarios and angles excluding identical views. We test the model on all gallery-probe pairs and report the mean over the probe angle and scenario as the accuracy of that specific setting. For brevity, we also report the accuracy on each scenario as the mean of the accuracies for all angles.

Table \ref{tab:psymo_gait_degraded} shows the performance of the GaitPT model on the 3 types of testing poses. In all scenarios of the testing set, the top recognition performance is obtained when utilizing poses from the fine-tuned artifact correction model in conjunction with the base pose estimation model. The poses obtained with the fine-tuned pose estimation model obtain a higher accuracy than those obtained with the frozen base model but a lower performance than our proposed approach. This indicates that our method improves the downstream recognition accuracy substantially.

We also analyse the impact of fine-tuning the models by testing the gait recognition accuracy on poses obtained on the original high quality videos. We utilize the same 3 approaches to obtain pose estimations, but without applying any video degradation to the frames. We employ the same evaluation protocol for gait recognition as in the previous experiment on highly degraded videos.

Table \ref{tab:psymo_gait_original} displays the performance of the GaitPT architecture on poses obtained from high quality images. The fine-tuned version of HRNet obtains the lowest overall performance with a mean recognition accuracy of 40.35\%. This shows that directly fine-tuning the pose estimation model on low quality data negatively impacts its performance on the high quality data as the original model yields an accuracy of 42.35\%. The improvement in performance on degraded videos comes at a cost of losing its generalization capabilities due to catastrophic forgetting. In contrast, the fine-tuned artifact correction model in combination with the original pose estimation model does not lose its generalization capabilities. FBCNN Fine-Tuned + HRNet Base manages to obtain a higher performance compared to both HRNet Base and HRNet Fine-Tuned, achieving an accuracy of 45.73\%. This result demonstrates the advantages of this 2-stage approach: the combination of models is more capable in estimating accurate poses in both low quality and high quality settings.

%% file: tables/pose_results.tex
\begin{table}[hbt!]
    \caption{Comparison of the pose estimation performance of the 3 experimental architectures in both low quality and high definition settings. The proposed approach obtains the highest Average Precision (AP) in both scenarios.}
    \label{tab:pose_results}
    \begin{center}
    \resizebox{\linewidth}{!}{
        \begin{tabular}{ l | c | c }
             \textbf{Dataset} & \textbf{Method} & \textbf{AP} \\
            \midrule
            \multirow{ 3}{*}{H264 compressed test set} 
            & HRNet Base & 0.783\\
            & HRNet Fine-Tuned & 0.935 \\
            & FBCNN Fine-Tuned + HRNet Base & \textbf{0.956} \\
            \midrule
            \multirow{ 3}{*}{Original test set}
            & HRNet Base & 0.935 \\
            & HRNet Fine-Tuned & 0.952 \\
            & FBCNN Fine-Tuned + HRNet Base & \textbf{0.967} \\
        \end{tabular}
    }
    \end{center}
\end{table}

%% file: tables/gait_results_degraded.tex
\begin{table*}[hbt!]
    \caption{Gait recognition accuracy comparison between the 3 types of poses obtained from \textbf{low quality videos}. The skeletons obtained with the fine-tuned artifact correction in conjunction with a frozen pose estimator yield the highest recognition performance.}
    \label{tab:psymo_gait_degraded}
    \begin{center}
    \resizebox{0.79\textwidth}{!}{
    \begin{tabular}{ c | l | cccccc | c }
         \textbf{Scenario} & \textbf{Test Poses} & 0$^{\circ}$  &    45$^{\circ}$  &    90$^{\circ}$  &    180$^{\circ}$  &    225$^{\circ}$  &    270$^{\circ}$ & \textbf{Mean} \\
        \midrule
        \multirow{ 3}{*}{NM (Normal)}  & HRNet Base & 32.67 & 31.33 & 32.67 & 28.67 & 35.33 & 33.33 & 32.33 \\
         & HRNet Fine-Tuned & 16.67 & 52.67 & 46.0 & 33.33 & 50.0 & 50.0 & 41.44 \\
         & FBCNN Fine-Tuned + HRNet Base & 42.67 & 50.0 & 46.67 & 46.0 & 50.67 & 48.0 & \textbf{47.33} \\
        
        \midrule
        \multirow{ 3}{*}{BG (Bag carry)}  & HRNet Base & 18.67 & 29.33 & 26.67 & 22.67 & 34.48 & 29.33 & 26.86 \\
         & HRNet Fine-Tuned & 20.67 & 36.67 & 40.67 & 28.67 & 41.38 & 43.33 & 35.23 \\
         & FBCNN Fine-Tuned + HRNet Base & 32.67 & 47.33 & 40.67 & 40.67 & 42.07 & 48.0 & \textbf{41.9} \\
        
        \midrule
        \multirow{ 3}{*}{CL (Clothing variation)}  & HRNet Base & 21.33 & 24.0 & 28.67 & 24.0 & 28.67 & 20.67 & 24.56 \\
         & HRNet Fine-Tuned & 18.67 & 20.67 & 23.33 & 25.33 & 36.67 & 28.67 & 25.56 \\
         & FBCNN Fine-Tuned + HRNet Base & 24.0 & 28.67 & 36.0 & 38.0 & 39.33 & 37.33 & \textbf{33.89} \\
        
        \midrule
        \multirow{ 3}{*}{WSS (Walking Speed Slow)}  & HRNet Base & 31.33 & 25.33 & 34.67 & 28.0 & 25.33 & 32.0 & 29.44 \\
         & HRNet Fine-Tuned & 22.67 & 48.0 & 46.0 & 33.33 & 38.67 & 46.0 & 39.11 \\
         & FBCNN Fine-Tuned + HRNet Base & 41.33 & 48.0 & 36.0 & 40.67 & 47.33 & 42.67 & \textbf{42.67} \\
        
        \midrule
        \multirow{ 3}{*}{WSF (Walking Speed Fast)}  & HRNet Base & 34.0 & 35.33 & 34.67 & 29.33 & 28.67 & 37.33 & 33.22 \\
         & HRNet Fine-Tuned & 28.0 & 42.67 & 40.67 & 31.33 & 41.33 & 46.0 & 38.33 \\
         & FBCNN Fine-Tuned + HRNet Base & 43.33 & 50.0 & 42.0 & 43.33 & 49.33 & 48.0 & \textbf{46.0} \\
        
        \midrule
        \multirow{ 3}{*}{PH (Phone while walking)}  & HRNet Base & 24.0 & 28.0 & 27.33 & 26.67 & 29.33 & 33.33 & 28.11 \\
         & HRNet Fine-Tuned & 25.33 & 44.0 & 41.33 & 25.33 & 32.67 & 37.33 & 34.33 \\
         & FBCNN Fine-Tuned + HRNet Base & 36.67 & 45.33 & 29.33 & 36.67 & 41.33 & 41.33 & \textbf{38.44} \\
        
        \midrule
        \multirow{ 3}{*}{TXT (Texting while walking)}  & HRNet Base & 30.0 & 25.33 & 30.67 & 25.33 & 22.67 & 24.67 & 26.44 \\
         & HRNet Fine-Tuned & 30.0 & 45.33 & 36.67 & 26.0 & 27.33 & 38.0 & 33.89 \\
         & FBCNN Fine-Tuned + HRNet Base & 38.0 & 42.67 & 33.33 & 34.0 & 32.0 & 36.67 & \textbf{36.11} \\

    \end{tabular}
    }
    \end{center}
\end{table*}

%% file: tables/gait_results_original.tex
\begin{table*}[hbt!]
    \caption{Gait recognition accuracy comparison between the 3 types of poses obtained from the \textbf{original high quality images}. Fine-tuning the HRNet on low-quality images decreases the performance on the high-quality data. However, the artifact correction model in conjunction with the original pose estimation model manage to obtain an increase in accuracy.}
    \label{tab:psymo_gait_original}
    \begin{center}
    \resizebox{0.79\textwidth}{!}{
    \begin{tabular}{ c | l | cccccc | c }
         \textbf{Scenario} & \textbf{Test Poses} & 0$^{\circ}$  &    45$^{\circ}$  &    90$^{\circ}$  &    180$^{\circ}$  &    225$^{\circ}$  &    270$^{\circ}$ & \textbf{Mean} \\
        \midrule
    \multirow{ 3}{*}{NM (Normal)}  & HRNet Base & 44.0 & 53.1 & 42.76 & 44.0 & 56.55 & 48.67 & 48.18 \\
     & HRNet Fine-Tuned & 36.0 & 58.62 & 54.48 & 37.33 & 56.55 & 52.67 & 49.28 \\
     & FBCNN Fine-Tuned + HRNet Base & 50.67 & 60.0 & 54.48 & 44.67 & 58.62 & 58.67 & \textbf{54.52} \\
    
    \midrule
    \multirow{ 3}{*}{BG (Bag carry)}  & HRNet Base & 39.33 & 46.0 & 30.67 & 41.33 & 48.97 & 40.0 & 41.05 \\
     & HRNet Fine-Tuned & 21.33 & 44.67 & 50.67 & 31.33 & 47.59 & 46.67 & 40.38 \\
     & FBCNN Fine-Tuned + HRNet Base & 38.67 & 44.0 & 40.0 & 44.0 & 50.34 & 53.33 & \textbf{45.06} \\
    
    \midrule
    \multirow{ 3}{*}{CL (Clothing variation)}  & HRNet Base & 25.33 & 35.33 & 36.0 & 39.33 & 54.0 & 38.67 & 38.11 \\
     & HRNet Fine-Tuned & 26.0 & 27.33 & 30.0 & 32.67 & 38.67 & 30.67 & 30.89 \\
     & FBCNN Fine-Tuned + HRNet Base & 29.33 & 34.67 & 36.0 & 38.67 & 44.67 & 45.33 & 38.11 \\
    
    \midrule
    \multirow{ 3}{*}{WSS (Walking Speed Slow)}  & HRNet Base & 42.67 & 47.33 & 43.33 & 46.0 & 47.33 & 50.0 & 46.11 \\
     & HRNet Fine-Tuned & 32.0 & 54.67 & 48.67 & 38.0 & 48.67 & 52.67 & 45.78 \\
     & FBCNN Fine-Tuned + HRNet Base & 43.33 & 55.33 & 45.33 & 40.67 & 52.67 & 58.0 & \textbf{49.22} \\
    
    \midrule
    \multirow{ 3}{*}{WSF (Walking Speed Fast)}  & HRNet Base & 50.0 & 50.0 & 40.67 & 39.33 & 48.67 & 50.67 & 46.56 \\
     & HRNet Fine-Tuned & 28.67 & 53.33 & 46.67 & 34.0 & 48.67 & 45.33 & 42.78 \\
     & FBCNN Fine-Tuned + HRNet Base & 44.67 & 49.33 & 37.33 & 44.67 & 56.67 & 52.67 & \textbf{47.56} \\
    
    \midrule
    \multirow{ 3}{*}{PH (Phone while walking)}  & HRNet Base & 39.33 & 48.0 & 36.67 & 43.33 & 42.0 & 32.67 & 40.33 \\
     & HRNet Fine-Tuned & 32.67 & 42.67 & 42.67 & 27.33 & 41.33 & 40.0 & 37.78 \\
     & FBCNN Fine-Tuned + HRNet Base & 49.33 & 53.33 & 32.0 & 42.0 & 48.0 & 46.67 & \textbf{45.22} \\
    
    \midrule
    \multirow{ 3}{*}{TXT (Texting while walking)}  & HRNet Base & 39.33 & 41.33 & 29.33 & 39.33 & 34.67 & 32.67 & 36.11 \\
     & HRNet Fine-Tuned & 30.67 & 44.67 & 40.67 & 26.67 & 38.0 & 32.67 & 35.56 \\
     & FBCNN Fine-Tuned + HRNet Base & 45.33 & 44.0 & 37.33 & 32.0 & 40.67 & 43.33 & \textbf{40.44} \\

    \end{tabular}
    }
    \end{center}
\end{table*}

%% file: sections/5.conclusions.tex
In this work, we propose a task-adapted video artifact correction model that is able to mitigate the effects of video compression on downstream pose-based gait recognition performance. Our model is guided by the supervisory signal of a frozen pose estimation model, and provides images that are purposely altered to obtain higher quality poses, irrespective of the underlying video quality. 

Furthermore, we analyzed two strategies for improving the performance of the human pose estimation and gait recognition tasks on highly compressed videos: fine-tuning HRNet, pretrained on the COCO keypoint detection dataset, and finetuning FBCNN, an artifact correction model, in combination with the pretrained HRNet model. We train and test these models on PsyMo, a gait analysis dataset with high quality videos of over 300 people walking. The results show that our proposed artifact correction pipeline increases the performance on both degraded and non-degraded videos. We showed that our approach does not lose generalization capabilities on high resolution video frames. 

Our work provides insights into the effects of video compression on downstream gait recognition performance and a simple and effective way to mitigate compression artifacts without using manually annotated data. We utilize ground truth poses from pretrained state-of-the-art models and show that they provide sufficient supervisory signal to train artifact correction models. As future work, we aim to extend our method to more general in-the-wild settings on gait datasets such as DenseGait \cite{cosma22gaitformer}, where we expect such an approach to have a significant impact due to the intrinsic low quality data.

%% file: sections/6.ack.tex
This work was partly supported by the NXP PhD Student Grants, by the Google IoT/Wearables Student Grants and by the Keysight Master Research Sponsorship.